\documentclass{article}

\usepackage{arxiv}

\usepackage[utf8]{inputenc} 
\usepackage[T1]{fontenc}    
\usepackage{hyperref}       
\usepackage{url}            
\usepackage{booktabs}       
\usepackage{amsfonts}       
\usepackage{nicefrac}       
\usepackage{microtype}      
\usepackage{lipsum}
\usepackage{graphicx}
\usepackage{xcolor}
\usepackage{natbib}
\usepackage{textcomp, gensymb} 
\graphicspath{ {./Figs/} }

\title{Comparative Study of UNet-based Architectures for Liver Tumor Segmentation in Multi-Phase Contrast-Enhanced Computed Tomography}

\author{
 Doan-Van-Anh Ly \\
  School of Computer Science and Engineering\\
  The Saigon International University\\
  Ho Chi Minh City 700000, Vietnam\\
  \texttt{anhldv1203@gmail.com} \\
   \And
 Thanh-Hai Le \\
  School of Computer Science and Engineering\\
  The Saigon International University\\
  Ho Chi Minh City 700000, Vietnam\\
  \texttt{lethanhhai@siu.edu.vn} \\
  \And
 Thi-Thu-Hien Pham \\
  School of Biomedical Engineering\\
  International University\\
  Ho Chi Minh City 700000, Vietnam\\
  Vietnam National University HCMC\\
  Ho Chi Minh City 700000, Vietnam\\
  \texttt{ptthien@hcmiu.edu.vn} \\
}

\begin{document}
\maketitle
\begin{abstract}
Segmentation of liver structures in multi-phase contrast-enhanced computed tomography (CECT) plays a crucial role in computer-aided diagnosis and treatment planning. In this study, we investigate the performance of UNet-based architectures for liver tumor segmentation, evaluating ResNet, Transformer-based, and State-space (Mamba) backbones initialized with pretrained weights. Our comparative analysis reveals that despite the theoretical advantages of modern architectures in modeling long-range dependencies, ResNet-based models demonstrated superior sample efficiency on this dataset. This suggests that the inherent inductive biases of Convolutional Neural Networks (CNNs) remain advantageous for generalizing on limited medical data compared to data-hungry alternatives. To further improve segmentation quality, we introduce attention mechanisms into the backbone, finding that the Convolutional Block Attention Module (CBAM) yields the optimal configuration. The ResNetUNet3+ with CBAM achieved the highest nominal performance with a Dice score of 0.755 and IoU of 0.662, while also delivering the most precise boundary delineation (lowest HD95 of 77.911). Critically, while statistical testing indicated that the improvement in mean Dice score was not significant (p > 0.05) compared to the baseline, the proposed model exhibited greater stability (lower standard deviation) and higher specificity (0.926). These findings demonstrate that classical ResNet architectures, when enhanced with modern attention modules, provide a robust and statistically comparable alternative to emerging methods, offering a stable direction for liver tumor segmentation in clinical practice.
\end{abstract}

\keywords{CBAM \and Grad-CAM \and Liver CECT \and Mamba \and ResNet \and Transformer \and Tumor Segmentation \and UNet3+}

\section{Introduction}
\label{sec:intro}
Liver cancer is among the leading causes of cancer-related mortality worldwide, and accurate delineation of the liver and its lesions in medical images is a prerequisite for computer-aided diagnosis, surgical planning, and treatment monitoring \cite{sung2021globocan, vietnamfact2022}. Contrast-enhanced computed tomography (CECT), particularly in multi-phase acquisition, provides detailed anatomical and functional information that makes it a preferred modality for liver imaging. However, manual annotation of liver structures is time-consuming, subject to inter-observer variability, and impractical for large-scale clinical deployment. This has motivated the development of several automated liver segmentation methods.

The application of deep learning has fundamentally transformed medical image analysis, with semantic segmentation being a critical prerequisite for efficient disease diagnosis and treatment by identifying organ or lesion pixels from images such as CT or MRI. Historically, this field has been dominated by Convolutional Neural Networks (CNNs). The foundational architecture in biomedical image segmentation is the UNet, known for its symmetric encoder-decoder structure and skip connections that enable the capture of both local and global contextual information for precise characterization of structures of interest \cite{yao2024fromcnn}. While UNet architectures have shown promising results, CNNs inherently struggle to model long-range dependencies (LRDs) due to the local nature of convolutional operations. To enhance performance, researchers have focused on modifying the UNet backbone by integrating elements that improve feature extraction and information flow. One significant development involves leveraging Residual Networks (ResNet), which utilize skip connections to prevent issues like vanishing gradients and performance degradation in deep networks, facilitating successful training and effective extraction of high-level features. Hybrid models like UNet-ResNet combine UNet’s structure with ResNet’s deep residual learning capabilities to capture complex features and contextual details, demonstrating superior segmentation accuracy compared to conventional Unet-based approaches in tasks such as liver tumor segmentation \cite{sheela2025enhancing}. This hybrid model was validated on a dataset consisting of 130 CT scans of liver cancers. Experimental results demonstrated impressive performance metrics for liver cancer segmentation, achieving an accuracy of 0.98 and a minimal loss of 0.10. Similarly, the UIGO model \cite{banerjee2025anovel} designed for automated medical image segmentation for liver tumor detection, merges the Unet architecture with Inception networks (specifically InceptionV3 blocks and an Inception-ResNet backbone) to enhance multi-scale feature extraction capabilities, addressing the challenge posed by the diversity in tumor shape, size, and texture. The model was tested using the LiTS, CHAOS, and 3D-IRCADb1 datasets for liver tumor detection. UIGO demonstrated exceptional results, achieving a segmentation accuracy of 99.93\%, a Dice Coefficient of 0.997, and an IoU of 0.998.

Further refinement has come through the integration of Attention Mechanisms. Models such as Attention UNet and AHCNet (Attention Hybrid Convolutional Network) integrate attention with skip connections to improve segmentation quality, enhancing the model’s capacity to identify fine features and focus on informative channels \cite{jiang2019ahcnet}. AHCNet was trained using 110 cases from the LiTS dataset (after removing the 3DIRCADb subset) and subsequently evaluated 20 cases in the 3DIRCADb dataset and 117 cases in a Clinical dataset. The proposed model achieved high performance in tumor segmentation accuracy, demonstrating an 11.6\% improvement in the dice global score compared to the baseline method on the 3DIRCADb dataset. The FSS-ULivR model designed for few-shot liver segmentation, employs improved attention gates, residual refinement, and multiscale skip connections in its decoder to restore spatial detail and generate accurate boundaries, selectively enhancing relevant features while suppressing irrelevant ones carried by conventional skip connections \cite{debnath2025fss}. This proposed model, trained on the LiTS dataset, was robustly validated on cross-datasets including 3DIRCADB01, CRLM, CT-ORG, and MSD-Task03-Liver. The model achieved an outstanding Dice coefficient of 98.94\%, an IoU of 97.44\%, and a specificity of 93.78\% on the LiTS test set, surpassing existing few-shot methods.

The realization that purely convolutional architectures were insufficient for capturing large-scale dependencies motivated the integration of the Transformer architecture, which excels at modeling global relationships through its self-attention mechanism. Networks like TransUNet \cite{chen2024transunet} and Swin-UNet \cite{cao2023swinunet} combine the hierarchical structure of UNet with Transformer modules in the encoder to improve long-range dependency modeling. However, a major drawback of Transformers in high-resolution and 3D medical image analysis is the high computational cost, as the self-attention mechanism scales quadratically with the input size. This limitation prompted research into State Space Models (SSMs), particularly the Mamba architecture, known for its ability to model long sequences with enhanced computational efficiency and linear scaling in feature size \cite{ma2024umamba}. Mamba-based models leverage the Visual State Space (VSS) block, which uses a Cross-Scan Module (CSM) to convert non-causal visual images into ordered patch sequences, thereby adapting Mamba to computer vision tasks \cite{wang2024mambaunet}. SegMamba \cite{xing2024segmamba} is introduced as the first method specifically utilizing Mamba for 3D medical image segmentation, featuring a tri-orientated Mamba (ToM) module to enhance sequential modeling of 3D features and a gated spatial convolution (GSC) module to retain spatial information. This model was tested on the new CRC-500 dataset, BraTS2023, and AIIB2023. On the BraTS2023 dataset, it achieved 93.61\%, 92.65\%, and 87.71\%, and HD95s of 3.37, 3.85, and 3.48 on WT, TC, and ET, respectively. VMAXL-UNet \cite{zhong2025visionmamba} fuses VSS blocks with extended LSTMs (xLSTM) within a UNet architecture to capture long-range dependencies while maintaining linear computational complexity. The model was tested on dermatological (ISIC17, ISIC18) and polyp segmentation (Kvasir-SEG, ClinicDB) datasets. The model significantly outperformed traditional CNNs and Transformers, achieving the best results with 90.1\% mIoU and 95.21\% DSC on the challenging ClinicDB dataset.

Recent advancements have focused on refining these hybrid architectures to better address specific limitations such as the semantic gap in feature fusion and boundary ambiguity. For instance, the MRC-TransUNet \cite{zhang2023noveldl} proposes replacing traditional skip connections with a Mobile and Residual Visual Transformer (MR-ViT) to bridge the semantic disparity between encoder and decoder, while employing a Reciprocal Attention (RPA) module to recover lost details. Similarly, CI-UNet \cite{zhang2024ciunet} integrates a ConvNeXt backbone with a novel Interoperability Attention (IOA) module, effectively capturing cross-dimensional interactions to enhance segmentation robustness in complex medical images. In the domain of dermatology, where lesion boundaries are often indistinct, an improved TransUNet framework \cite{wang2025deepsl} was developed that fuses a Transformer for global context with a convolutional branch for local texture, utilizing a Boundary-Guided Attention (BGA) mechanism to explicitly refine edge reconstruction. Beyond segmentation, hybrid models have also shown utility in classification tasks; the Meningioma Feature Extraction Model (MFEM) \cite{zhang2024dlradiomics} combines ConvNeXt and Transformer layers to extract deep features which, when integrated with radiomics data from peritumoral edema regions, significantly improve the preoperative grading of meningiomas.

Recent efforts to improve liver segmentation from multi-phase CECT images have focused on building standardized datasets and developing advanced deep learning models. Luo et al. \cite{lou2025comprehensive} published a comprehensive 3D multi-phase CECT dataset for primary liver cancer, covering three major types: HCC, ICC, and cHCC-CCA. The dataset includes 278 cancer cases and 83 non-cancer cases, with over 50,000 annotated slices, providing a valuable foundation for training and evaluating classification and segmentation models. Meanwhile, Hu et al. \cite{hu2024trustworthy} proposed the TMPLiTS framework for reliable multi-phase liver tumor segmentation, incorporating Dempster–Shafer theory to model uncertainty and a MEMS expert-mixing mechanism to fuse information across phases. Their method outperformed existing approaches in both accuracy and reliability, especially under noisy or incomplete data conditions. Additionally, Nayantara et al. \cite{Nayantara2024automatic} developed an automatic liver segmentation model using an enhanced SegNet architecture combined with an ASPP module and leaky ReLU activation. The model achieved Dice scores above 96\% in the portal venous phase and over 93\% in other phases (arterial, delayed and plain CT phases), demonstrating superior performance and strong clinical applicability

Furthermore, the general opacity of sophisticated deep learning models (often operating as "black boxes") raises concerns for clinical practitioners regarding the rationale behind segmentation outputs, necessitating greater Explainable Artificial Intelligence (xAI). The field of xAI has largely focused on image classification, but methods are being extended to computer vision domains like image segmentation, specifically in medical image analysis \cite{lamprou2024evaluation}. Seg-Grad-CAM \cite{vinogradova2020towards} is one such extension of the popular Grad-CAM algorithm applied to segmentation. However, a key nuance associated with its utilization is that Seg-Grad-CAM assigns a single coefficient to each activation map after global average pooling of the gradient matrix, meaning it does not incorporate spatial considerations when generating explanations for specific regions within a segmentation map. To solve this, Seg-XRes-CAM \cite{hasany2023segxrescam} was proposed, taking inspiration from HiResCAM \cite{draelos2020hirescam}, to generate location-aware and spatially localized explanations for image segmentation regions, demonstrating a higher degree of visual agreement with model-agnostic methods like RISE compared to Seg-Grad-CAM.

This study employs the Primary Liver Cancer CECT Imaging Dataset [14], a well-curated resource obtained from a single medical facility. The dataset offers comprehensive 3D CECT scans with expert annotations delineating both liver and lesion regions, serving as a solid foundation for the development and validation of advanced segmentation models. The methodological approach begins with an assessment of various UNet and UNet3+ architectures. Upon selection of the most effective baseline model, a detailed investigation is subsequently performed by incorporating four distinct module variants—SE, CBAM, ASPP, and a combined CBAM-ASPP configuration—to identify the optimal architecture for precise lesion segmentation.

\section{Materials and Methods}
\label{sec:materials}
\subsection{Dataset}
The Primary Liver Cancer CECT Imaging Dataset \cite{lou2025comprehensive} contains 278 liver cancer cases and 83 non-liver cancer cases, each with full 3D multi-phase contrast-enhanced CT (CECT) scans (Plain, Arterial, Venous, and Delayed phases). Both the liver and lesion regions were annotated in NIfTI format. 

\begin{figure}[htb!] 
    \centering
    \includegraphics[width=0.8\textwidth]{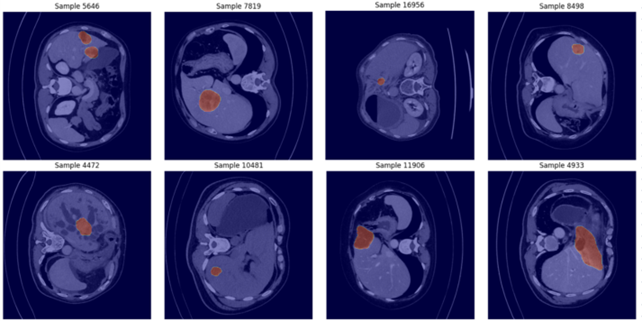}
    \caption{CT slice samples with ground truth.}
    \label{fig:fig1}
\end{figure}

Preprocessing was necessary to prepare the dataset, which consisted of 3D volumes, for training. A verification pipeline was developed to ensure consistency between CT volumes and their corresponding segmentation masks of tumor regions (ground truth), as shown in Fig. \ref{fig:fig1}. Each CT–mask pair underwent validation for dimensional alignment, slice count, file integrity, and the presence of annotated lesions. The pairs identified as invalid or mismatched were excluded from further analysis. The results of this verification process were documented in a metadata CSV file, which subsequently served as an index for all experimental procedures. This preprocessing pipeline produced a reliable and balanced dataset of 2D liver tumor slices, enabling efficient experimentation with different segmentation architectures.

\begin{table}[htb!]
 \caption{Distribution of training, validation and testing sets after preprocessing 3D CECT scans.}
  \centering
  \begin{tabular}{llll}
    \toprule
    \textbf{Dataset}     & \textbf{Train}     & \textbf{Val} & \textbf{Test} \\
    \midrule
    Primary Liver Cancer CECT Imaging Dataset & 5335  & 3180 & 2211     \\
    \bottomrule
  \end{tabular}
  \label{tab:table1}
\end{table}

For model training, 2D slices were extracted from the valid CT–mask pairs. Each slice was normalized to zero mean and unit variance, then resized. To enhance model generalization, data augmentation was applied during training, consisting of random horizontal and vertical flips, as well as 90\degree rotations. For each patient, the data was divided by phase, allocating two phases for training, one for validation, and one for testing.  Table \ref{tab:table1} presents the number of slices in each set following validation of the CT-mask pairs from the dataset.

\begin{figure}[htb!] 
    \centering
    \includegraphics[width=0.8\textwidth]{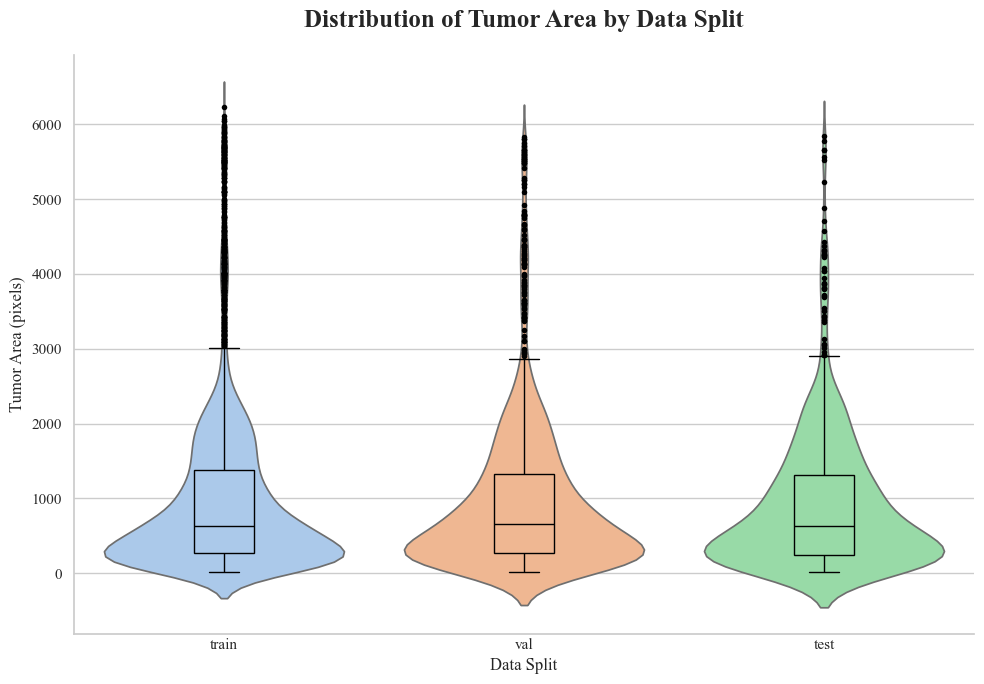}
    \caption{Distribution of training, validation and testing sets on tumor area.}
    \label{fig:fig2}
\end{figure}

Figure \ref{fig:fig2} illustrates the distribution of the dataset across three subsets. For all data splits, the distribution is heavily concentrated near zero, with the 25th percentile and median pixel areas being very low. To process this data, images were standardized to a resolution of \(256\times256\) pixels. While downsampling presents inherent challenges for such small objects, this resolution was selected to optimize the trade-off between spatial detail and computational efficiency, significantly reducing memory overhead while preserving essential semantic features for the network.

\subsection{Models}
\subsubsection{Base Model Selection}
In this study, two widely used segmentation architectures were selected as the foundation for experimentation: UNet \cite{ronneberger2015unet} and UNet3+ \cite{huang2020unet3p}.

The UNet architecture has become the de facto standard for medical image segmentation tasks due to its encoder–decoder design with skip connections. The encoder captures hierarchical features through successive convolution and pooling operations, while the decoder progressively reconstructs spatial details. The skip connections ensure that fine-grained localization is preserved, making UNet particularly effective for segmenting small and irregular structures such as liver tumors.

Building upon UNet, UNet3+ introduces full-scale skip connections that aggregate feature maps from multiple encoder and decoder levels. This design allows for a more comprehensive fusion of semantic and spatial information, addressing the limitations of conventional UNet in balancing detail preservation with global contextual understanding. In liver tumor segmentation, this capability is especially beneficial since tumors may vary significantly in size, shape, and intensity.

To further enhance feature extraction, both architectures were integrated with different backbone networks as encoders. Backbones such as Mamba, ResNet, … provide pre-trained feature representations that can accelerate convergence and improve generalization. Incorporating these backbones allows the models to leverage knowledge from large-scale natural image datasets, which is particularly advantageous given the limited size of medical imaging datasets.

\begin{figure}[htb!] 
    \centering
    \includegraphics[width=0.8\textwidth]{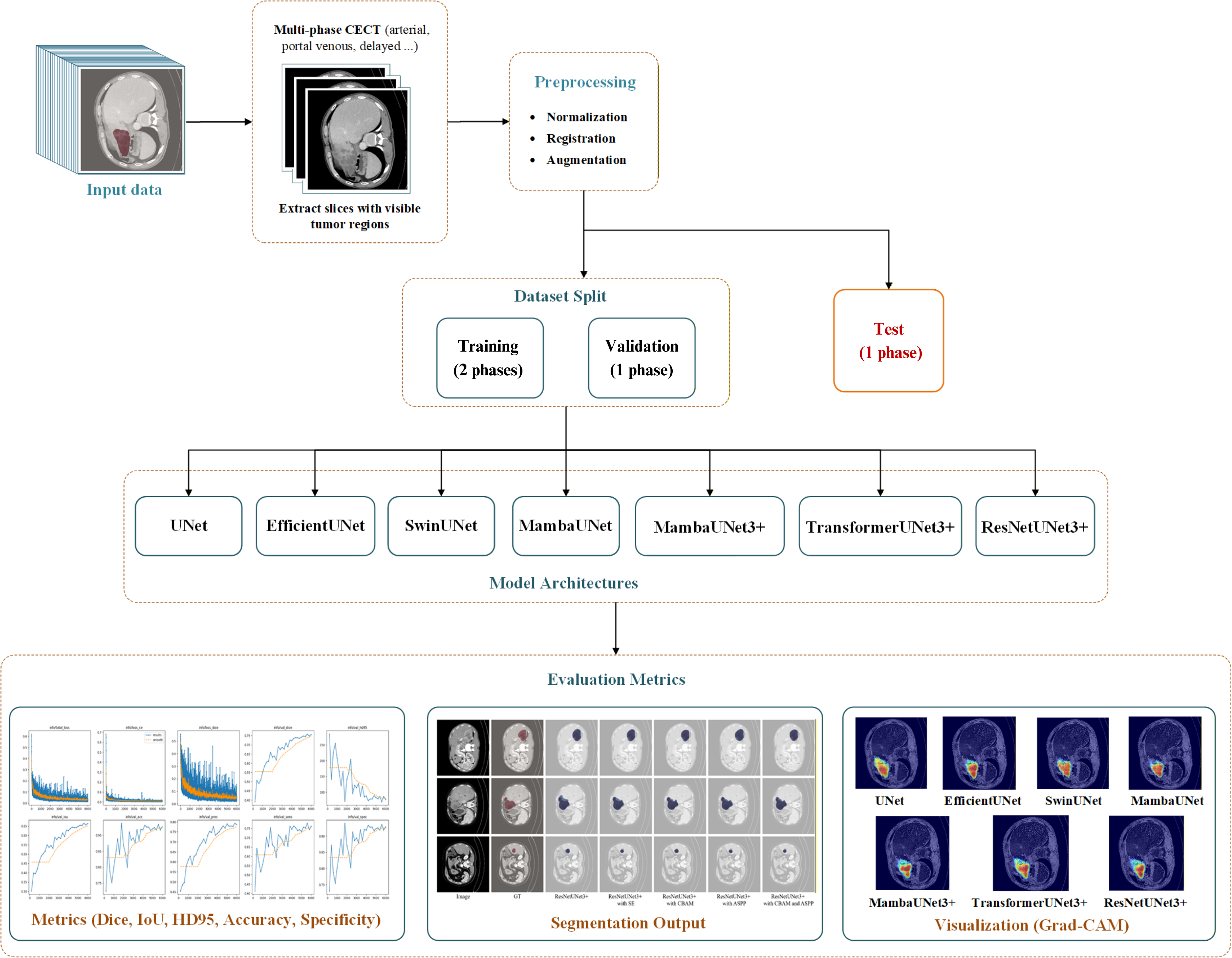}
    \caption{Research framework.}
    \label{fig:fig3}
\end{figure}

By comparing the performance of UNet and UNet3+ under various backbone configurations, this study aims to identify a balance between segmentation accuracy, boundary precision, and computational efficiency in the context of liver tumor segmentation from CECT scans. Figure \ref{fig:fig3} presents the operational framework that will be utilized throughout this study.

\subsubsection{ResNet UNet3+ with attention mechanisms}
The best-performing architecture in this study was based on the combination of a ResNet \cite{he2016deepresidual} encoder backbone and a UNet3+ decoder, enhanced by the Convolutional Block Attention Module (CBAM) \cite{woo2018cbam, lieu2025enhanced}. This section provides a detailed rationale for the choice of each component and their synergistic effect on liver tumor segmentation from CECT images.

\begin{figure}[htb!] 
    \centering
    \includegraphics[width=0.8\textwidth]{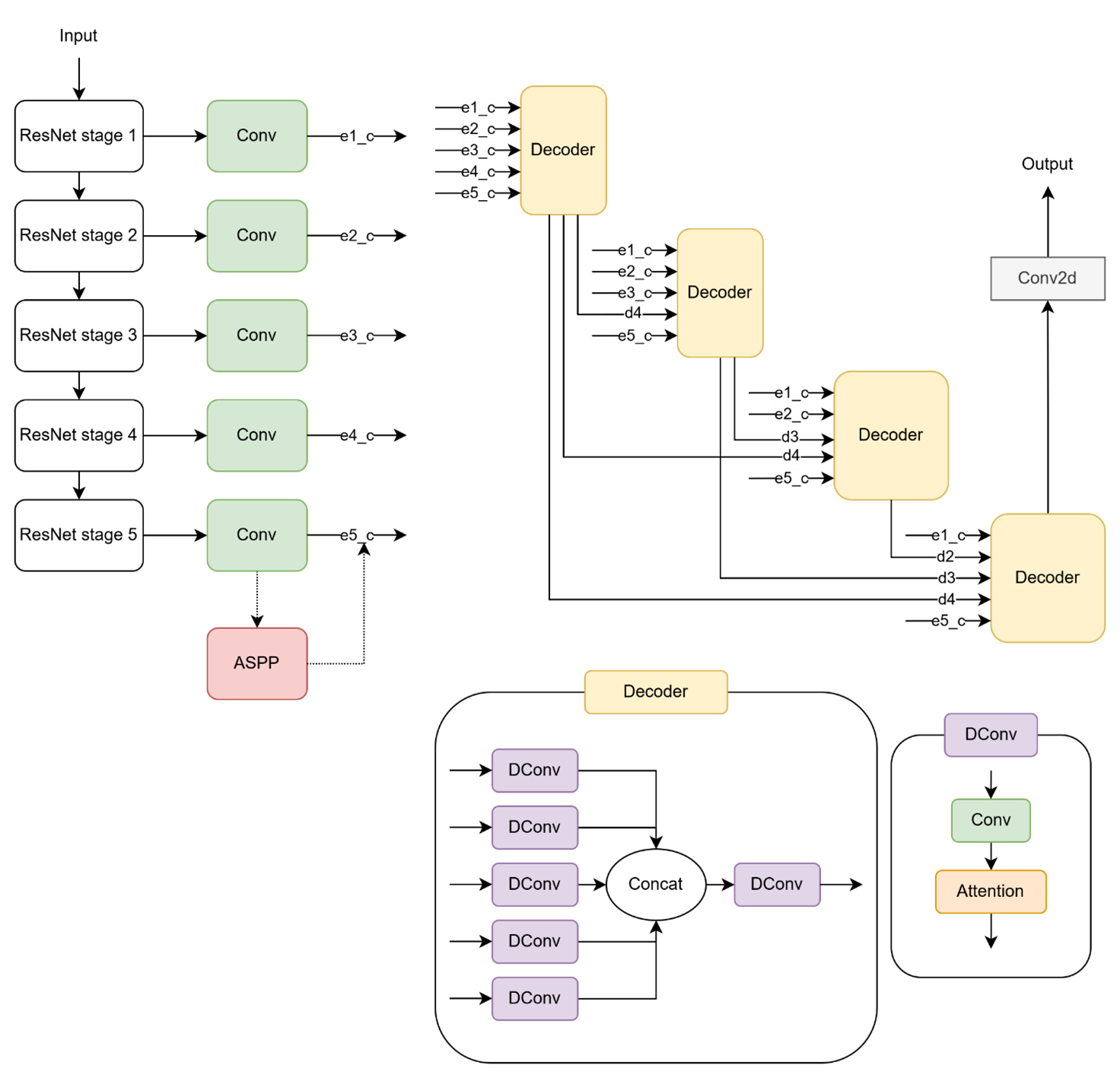}
    \caption{Illustration of ResNetUNet3+ variant.}
    \label{fig:fig4}
\end{figure}

The encoder network plays a critical role in extracting hierarchical features from the input images. ResNet was selected as the backbone due to its residual learning framework, which facilitates the training of very deep networks by introducing skip connections that mitigate the vanishing gradient problem. This allows the model to capture both low-level spatial details and high-level semantic features. Figure \ref{fig:fig4} presents the architecture of ResNetUNet3+ incorporating the ASPP module, illustrating one of the variants examined in this study.

In medical imaging, where tumor boundaries are often subtle and tumor appearance varies across phases (plain, arterial, venous, delayed), ResNet provides a balance of computational efficiency and robust feature extraction. Unlike lightweight networks that may underfit complex structures, or Transformer-based encoders that require larger datasets and computational resources, ResNet50 has demonstrated stable performance in transfer learning scenarios with limited annotated data. Its pre-training on ImageNet further supports generalization by offering strong initialization for low-level edge, texture, and shape features, which are essential for distinguishing lesions from healthy liver tissue.

While the original UNet has proven highly effective for biomedical segmentation tasks, it relies on single-scale skip connections that may not fully capture the wide variation in lesion size. UNet3+ extends this architecture by introducing full-scale skip connections and a redesigned decoder path. Specifically, UNet3+ aggregates feature maps from encoder and decoder layers at multiple scales, ensuring that fine-grained details (e.g., small lesion edges) are combined with high-level semantic context (e.g., overall tumor shape).

For liver tumor segmentation, this design is particularly advantageous. Tumors can vary significantly in size and intensity across patients and CECT phases; full-scale feature fusion enables the network to remain sensitive to both microscopic nodules and large heterogeneous masses. Additionally, the redesigned decoder mitigates the semantic gap between encoder and decoder features, leading to more precise reconstruction of lesion boundaries.

Although UNet3+ with ResNet provides strong baseline performance, not all extracted features are equally relevant for tumor segmentation. To address this, attention mechanisms were integrated into the decoder to enhance feature discriminability. Two approaches were considered:
\begin{itemize}
\item Squeeze-and-Excitation (SE) \cite{hu2018se}: SE modules perform global average pooling followed by channel-wise recalibration, allowing the network to assign higher weights to informative feature channels. While effective in many applications, SE does not explicitly model spatial dependencies.
\item Convolutional Block Attention Module (CBAM): CBAM extends the SE concept by sequentially applying channel attention and spatial attention. Channel attention identifies which feature maps are most relevant (e.g., texture-sensitive channels for tumor edges), while spatial attention highlights discriminative regions within each feature map (e.g., the tumor area itself rather than its surroundings).
\end{itemize}

By jointly refining both channel and spatial information, CBAM directs the network’s focus toward tumor-relevant signals across phases while suppressing background noise such as blood vessels or liver parenchyma. This dual attention is critical in CECT imaging, where tumors may exhibit subtle intensity differences relative to the background.

To enhance the interpretability of the segmentation models, Gradient-weighted Class Activation Mapping (Grad-CAM) \cite{selvaraju2019gradcam, gildenblat2021pytorchcam} was applied. The last convolutional layer of each network was identified as the target for backpropagation-based saliency generation. During inference, the predicted segmentation mask was used to define the target category (foreground vs. background). Grad-CAM was then computed by propagating gradients from the target class back through the final convolutional feature maps, producing a class-specific heatmap. These heatmaps were normalized and overlaid on the original CT images, enabling qualitative assessment of whether the model focused on clinically relevant tumor regions when making predictions.

\subsection{Evaluation Metrics}
In this study, both region-based and boundary-based metrics were employed to evaluate segmentation performance. Each metric captures different aspects of model effectiveness, from volumetric overlap to boundary accuracy. The following notations are used:
\begin{itemize}
    \item \textit{P}: set of pixels (or voxels) predicted as lesion (positive region).
    \item \textit{G}: set of pixels (or voxels) belonging to the ground truth lesion.
    \item \textit{TP} (True Positives): pixels correctly predicted as lesion.
	\item \textit{FP} (False Positives): pixels incorrectly predicted as lesion.
	\item \textit{FN} (False Negatives): pixels belonging to lesion but missed by the model.
	\item \textit{TN} (True Negatives): pixels correctly predicted as background.
    \item \textit{d(p,g)}: Euclidean distance between a predicted boundary point \textit{p} and a ground truth boundary point \textit{g}.
\end{itemize}

\textbf{Dice Similarity Coefficient (DSC)} evaluates the overlap between prediction and ground truth. Here, \(\mid\textit{P}\cap\textit{G}\mid\) represents the number of pixels correctly predicted as lesion (TP), while \(\mid\textit{P}\mid\) and \(\mid\textit{G}\mid\) represent the total number of predicted and true lesion pixels, respectively. DSC balances false positives and false negatives, making it one of the most reliable metrics for medical image segmentation. A higher DSC indicates a stronger agreement with the reference annotation.

\begin{equation}\label{eq1}
   {DSC} = \frac{2\mid P \cap G\mid}{\mid P\mid + \mid G \mid} 
\end{equation}

\textbf{Intersection over Union (IoU)} measures the ratio of overlap between predicted and ground truth lesion areas to their combined area. The denominator \(\mid\textit{P}\cup\textit{G}\mid\) corresponds to all pixels labeled as lesion either by the model or in the ground truth. IoU provides a stricter evaluation compared to DSC, as even small mismatches reduce the score.

\begin{equation}\label{eq2}
   {IoU} = \frac{\mid P \cap G\mid}{\mid P \cup G \mid} 
\end{equation}

\textbf{Hausdorff Distance (95\%) (HD95)} evaluates the similarity of the predicted and ground truth boundaries. For each boundary point p in the prediction, the minimum distance to any ground truth point g is computed, and vice versa. The 95th percentile is used instead of the maximum to reduce the effect of outliers caused by noise or annotation errors. A lower HD95 value indicates that the model’s predicted boundary closely follows the true tumor contour, which is crucial for clinical applications.

\textbf{Accuracy} measures the proportion of correctly classified pixels relative to all pixels. While simple to interpret, accuracy can be misleading in medical segmentation where lesion regions are small compared to the background, as models could achieve high accuracy simply by predicting background.

\begin{equation}\label{eq3}
   {Accuracy} = \frac{TP + TN}{TP + TN + FP + FN}
\end{equation}

\textbf{Precision} indicates the proportion of predicted lesion pixels that are correct. TP represents correctly segmented lesion pixels, while FP counts pixels wrongly identified as lesions. High precision reflects the model’s ability to avoid false positives, ensuring that detected tumor regions are likely real.

\begin{equation}\label{eq4}
   {Precision} = \frac{TP}{TP + FP}
\end{equation}

\textbf{Sensitivity} measures the proportion of actual lesion pixels correctly identified. A high value means the model rarely misses tumor regions, though it may include some background (false positives). In clinical contexts, high sensitivity is essential to avoid missing critical lesions.
	
\begin{equation}\label{eq5}
   {Sensitivity} = \frac{TP}{TP + FN}
\end{equation}

\textbf{Specificity} quantifies the ability of the model to correctly classify background pixels. High specificity means healthy tissues are not mistakenly identified as tumor. This reduces the risk of over-segmentation, which could otherwise lead to unnecessary clinical concern.

\begin{equation}\label{eq6}
   {Specificity} = \frac{TN}{TN + FP}
\end{equation}

By combining these complementary metrics, the evaluation framework balances volumetric accuracy (DSC, IoU, Accuracy), boundary alignment (HD95), and classification behavior (Precision, Sensitivity, Specificity). This ensures both quantitative rigor and clinical interpretability of the segmentation results.

\section{Results and Discussion}
\label{sec:results}
\subsection{Results}
\subsubsection{Experiment setup}
All segmentation experiments were performed within the Kaggle computing environment to ensure reproducibility. Detailed specifications for the hardware and training hyperparameters are provided in Tables \ref{tab:table2} and \ref{tab:table3}, respectively. The training regimen for all models, which include selected data augmentation techniques, utilized a consistent set of hyperparameters. It is noteworthy that while most models, including the best-performing architecture, used an input image size of \(256\times256\) pixels (as indicated in Table \ref{tab:table2}), models incorporating a Transformer-based backbone required a smaller input resolution of \(224\times224\) pixels due to architectural constraints.

\begin{table}[htb!]
 \caption{Kaggle experimental environment.}
  \centering
  \begin{tabular}{ll}
    \toprule    
    \textbf{Processor}               & Intel(R)Xeon(R)CPU@2.20GHz    \\
    \textbf{RAM}                     & 31.4GB    \\
    \textbf{Graphics card}           & Tesla T4 GPU    \\
    \textbf{Programming language}    & Python 3.11.11    \\
    \textbf{Deep learning framework} & Pytorch 2.6.0+cu124    \\
    \bottomrule
  \end{tabular}
  \label{tab:table2}
\end{table}

\begin{table}[htb!]
 \caption{Kaggle experimental environment.}
  \centering
  \begin{tabular}{llll}
    \toprule    
    \textbf{Hyperparameter} & \textbf{Value} & \textbf{Hyperparameter} & \textbf{Value}\\    
    \midrule
    Initial learning rate & 0.01  & Optimizer & SGD \\
    Batch size & 4  & Epoch & 100 \\
    Image size & \(256\times256\)  & Momentum & 0.9 \\
    Max iterations & 6000  & Weight decay & 0.0001 \\
    \bottomrule
  \end{tabular}
  \label{tab:table3}
\end{table}

\begin{figure}[htb!] 
    \centering
    \includegraphics[width=0.8\textwidth]{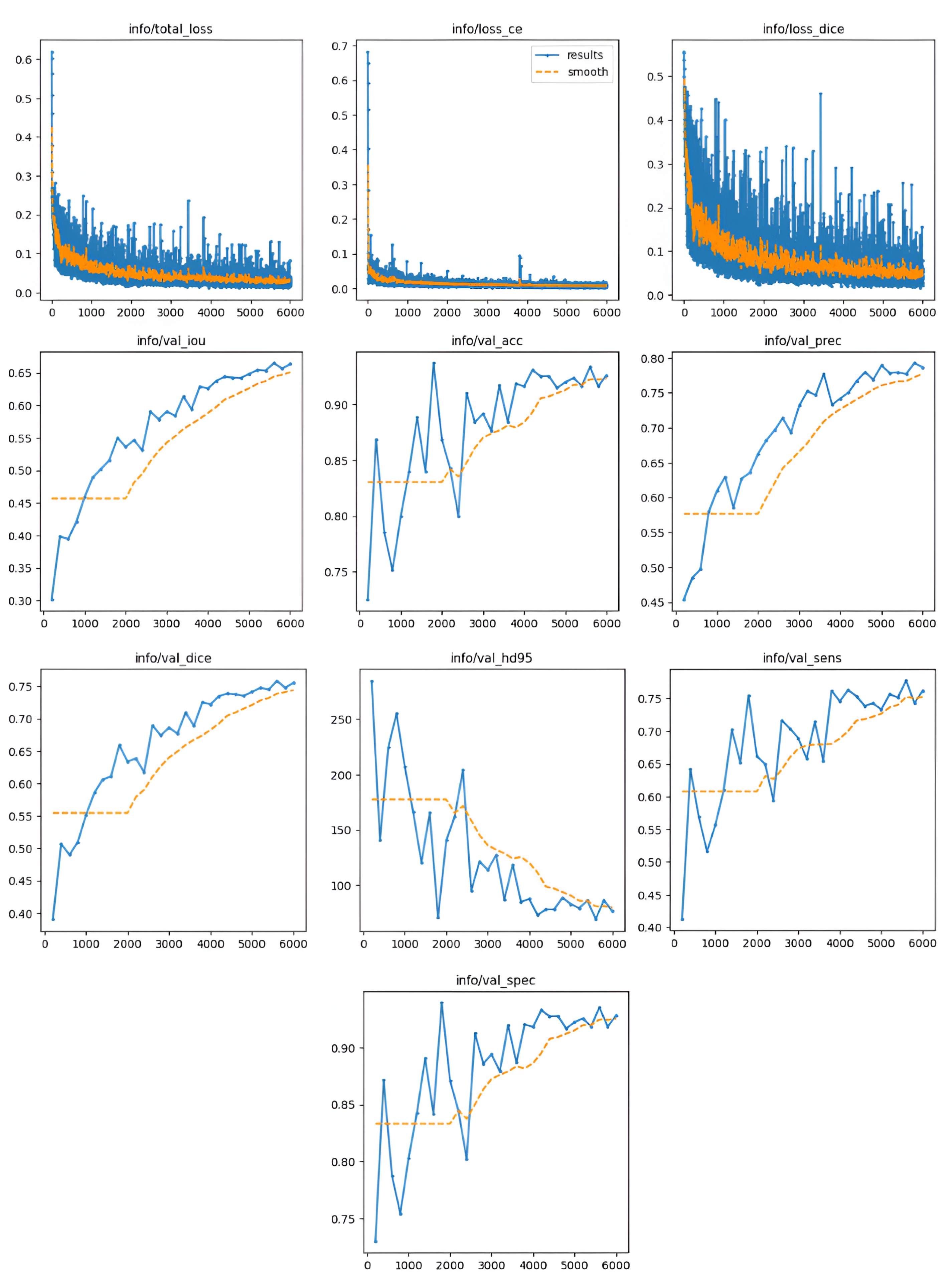}
    \caption{Training and validation process visualization of proposed model.}
    \label{fig:fig5}
\end{figure}

The learning efficacy and stability of the model are visually documented in Fig. \ref{fig:fig5}, which plots key training and validation metrics across all optimization steps. The graphs confirm that the model achieved stable convergence, as evidenced by the rapid initial decline and subsequent low, stable values across all loss functions (total\_loss, loss\_ce, loss\_dice). Importantly, the primary validation metrics, including the Dice coefficient (val\_dice) and Intersection over Union (val\_iou), show a continuous, robust improvement throughout the training duration. This simultaneous behavior substantiates the model's ability to learn the CECT segmentation task effectively while maintaining strong generalization capability on the validation dataset.

\subsubsection{The UNet and UNet3+ with different backbones}
To evaluate the effectiveness of the proposed architecture, multiple segmentation models were trained and compared on the Primary Liver Cancer CECT dataset using the evaluation metrics described earlier. Table \ref{tab:table4} summarizes the performance of all baseline and enhanced models.

The experiments include:
\begin{itemize}
    \item UNet based models: UNet, EfficientUNet, SwinUNet, MambaUNet.
    \item UNet3+ based models: MambaUNet3+, ResNetUNet3+, TransformerUNet3+.
\end{itemize}

\begin{table}[htb!]
 \caption{Performance of models on the testing set. (Note that highest and lowest scores are highlighted in blue and red fonts, respectively.)}
  \centering
  \begin{tabular}{lllllllll}
    \toprule    
    \textbf{Model} & \textbf{Param} & \textbf{Dice(DSC)} & \textbf{HD95} & \textbf{IoU} & \textbf{Acc} & \textbf{Pre} & \textbf{Sen} & \textbf{Spe}   \\    
    \midrule
    UNet & 1.8M  & \textbf{\textcolor{red}{0.506}} & \textbf{\textcolor{red}{224.710}} & \textbf{\textcolor{red}{0.491}} & \textbf{\textcolor{red}{0.786}} & \textbf{\textcolor{red}{0.563}} & \textbf{\textcolor{red}{0.532}} & \textbf{\textcolor{red}{0.789}} \\
    EfficientUNet & 13.2M & 0.601 & 165.982 & 0.504 & 0.838 & 0.622 & 0.649 & 0.840 \\
    SwinUNet & 27.6M & 0.617 & 129.808 & 0.512 & 0.874 & 0.610 & 0.678 & 0.876 \\
    MambaUNet & 19.1M & 0.646 & 130.916 & 0.548 & 0.875 & 0.641 & 0.705 & 0.877 \\
    MambaUNet3+ & 36.1M & 0.710 & 112.901 & 0.617 & 0.890 & 0.735 & 0.724 & 0.892 \\
    TransformerUNet3+ & 53.3M & 0.656 & 159.379 & 0.565 & 0.843 & 0.677 & 0.678 & 0.845 \\
    ResNetUNet3+ & 31.1M & \textbf{\textcolor{blue}{0.746}} & \textbf{\textcolor{blue}{88.082}} & \textbf{\textcolor{blue}{0.657}} & \textbf{\textcolor{blue}{0.915}} & \textbf{\textcolor{blue}{0.764}} & \textbf{\textcolor{blue}{0.768}} & \textbf{\textcolor{blue}{0.916}} \\ \bottomrule
  \end{tabular}
  \label{tab:table4}
\end{table}

The original UNet achieved a Dice coefficient of 0.506, an IoU of 0.419, and a relatively high HD95 of 224.71, reflecting its limited ability to capture complex lesion boundaries in contrast-enhanced CT images. Although this baseline provided a useful point of comparison, its performance highlighted the need for more powerful architectures capable of modeling richer spatial and contextual information.

Replacing the baseline with more recent backbones demonstrated varying levels of improvement. EfficientUNet and SwinUNet achieved Dice scores of 0.601 and 0.617, respectively, showing that both convolutional and transformer-based encoders enhanced segmentation accuracy over the vanilla UNet. The MambaUNet further improved performance with a Dice of 0.646 and reduced HD95, suggesting that state-space modeling could effectively capture long-range dependencies. Notably, the MambaUNet3+ variant, which integrated the multi-scale full-scale skip connections of UNet3+, produced a substantial leap in accuracy with a Dice score of 0.710 and IoU of 0.617, demonstrating the strength of this hybrid design.

Among the tested variants, the ResNetUNet3+ consistently outperformed other backbones. With a Dice score of 0.746, IoU of 0.657, and the lowest HD95 of 88.08 at this stage, this model provided a strong balance between segmentation accuracy and boundary precision. The residual connections within ResNet, combined with the dense feature aggregation of UNet3+, appeared to enhance both low- and high-level feature representation, leading to improved lesion delineation.

\begin{figure}[htb!] 
    \centering
    \includegraphics[width=0.8\textwidth]{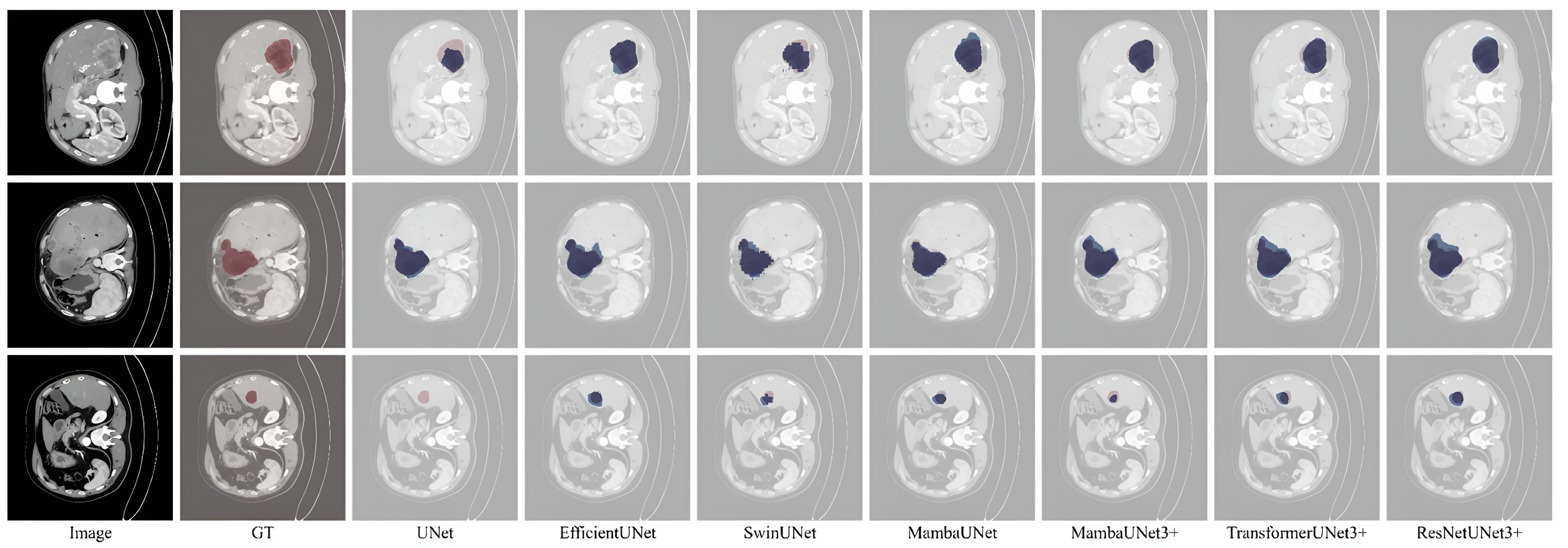}
    \caption{The visual comparison of seven segmentation methods against ground truth on the testing set. The predicted and GT are highlighted in blue and red, respectively.}
    \label{fig:fig6}
\end{figure}

Predicted masks from the best-performing model, ResNetUNet3+, show more accurate tumor boundary delineation and fewer false positives compared to other architectures. Example cases illustrate both successful segmentations and challenging scenarios, such as small lesions or lesions adjacent to vessels, where performance remained limited. Visual comparisons highlight that ResNetUNet3+ with medium size of parameters (31.1 million) is more robust in capturing tumors than baseline UNet or Mamba- and Transformer-based alternatives as shown in Fig. \ref{fig:fig6}.

\subsubsection{Ablation study}
Building on the ResNetUNet3+ baseline, we conducted an ablation study to evaluate the impact of different attention mechanisms, including Squeeze-and-Excitation (SE), Convolutional Block Attention Module (CBAM), and Atrous Spatial Pyramid Pooling (ASPP).

\begin{table}[htb!]
 \caption{Performance of ResNet Unet3+ variants on the testing set. (Note that highest scores are highlighted in blue fonts.)}
  \centering
  \begin{tabular}{|p{3cm}|p{1cm}|p{1cm}|p{1cm}|p{1cm}|p{1cm}|p{1cm}|p{1cm}|p{1cm}|p{1cm}|}   
    \toprule    
    \textbf{Model} & \textbf{Param} & \textbf{Dice (DSC)} & \textbf{HD95} & \textbf{IoU} & \textbf{Acc} & \textbf{Pre} & \textbf{Sen} & \textbf{Spe} & \textbf{P-Value}   \\    
    \midrule
    ResNetUNet3+ & 31.1M & \(0.746 \pm 0.284\) & \(6.70 \pm 9.13\) & \(0.657 \pm 0.275\) & \(0.915 \pm 0.274\) & \(0.764 \pm 0.287\) & \(0.768 \pm 0.302\) & \(0.916 \pm 0.274\) & - \\
    ResNetUNet3+ with SE & 31.2M & \(0.747 \pm 0.288\) & \(6.51 \pm 8.51\) & \(0.659 \pm 0.276\) & \(0.903 \pm 0.291\) & \(0.767 \pm 0.289\) & \(0.761 \pm 0.306\) & \(0.904 \pm 0.292\) & 0.1448 \\
    ResNetUNet3+ with CBAM & 31.2M & \textbf{\textcolor{blue}{\(0.755 \pm 0.268\)}} & \(6.70 \pm 10.01\) & \textbf{\textcolor{blue}{\(0.662 \pm 0.261\)}} & \textbf{\textcolor{blue}{\(0.925 \pm 0.257\)}} & \(0.768 \pm 0.273\) & \textbf{\textcolor{blue}{\(0.777 \pm 0.288\)}} & \textbf{\textcolor{blue}{\(0.926 \pm 0.258\)}} & 0.2872 \\
    ResNetUNet3+ with ASPP & 31.3M & \(0.735 \pm 0.295\) & \(7.19 \pm 10.41\) & \(0.645 \pm 0.282\) & \(0.897 \pm 0.299\) & \(0.756 \pm 0.298\) & \(0.754 \pm 0.312\) & \(0.898 \pm 0.299\) & 0.0005 \\
    ResNetUNet3+ with CBAM and ASPP & 31.4M & \(0.753 \pm 0.275\) & \textbf{\textcolor{blue}{\(6.11 \pm 7.89\)}} & \(0.662 \pm 0.265\) & \(0.912 \pm 0.278\) & \textbf{\textcolor{blue}{\(0.777 \pm 0.281\)}} & \(0.761 \pm 0.291\) & \(0.914 \pm 0.278\) & 0.6923 \\
    \bottomrule
  \end{tabular}
  \label{tab:table5}
\end{table}

\begin{figure}[htb!] 
    \centering
    \includegraphics[width=0.8\textwidth]{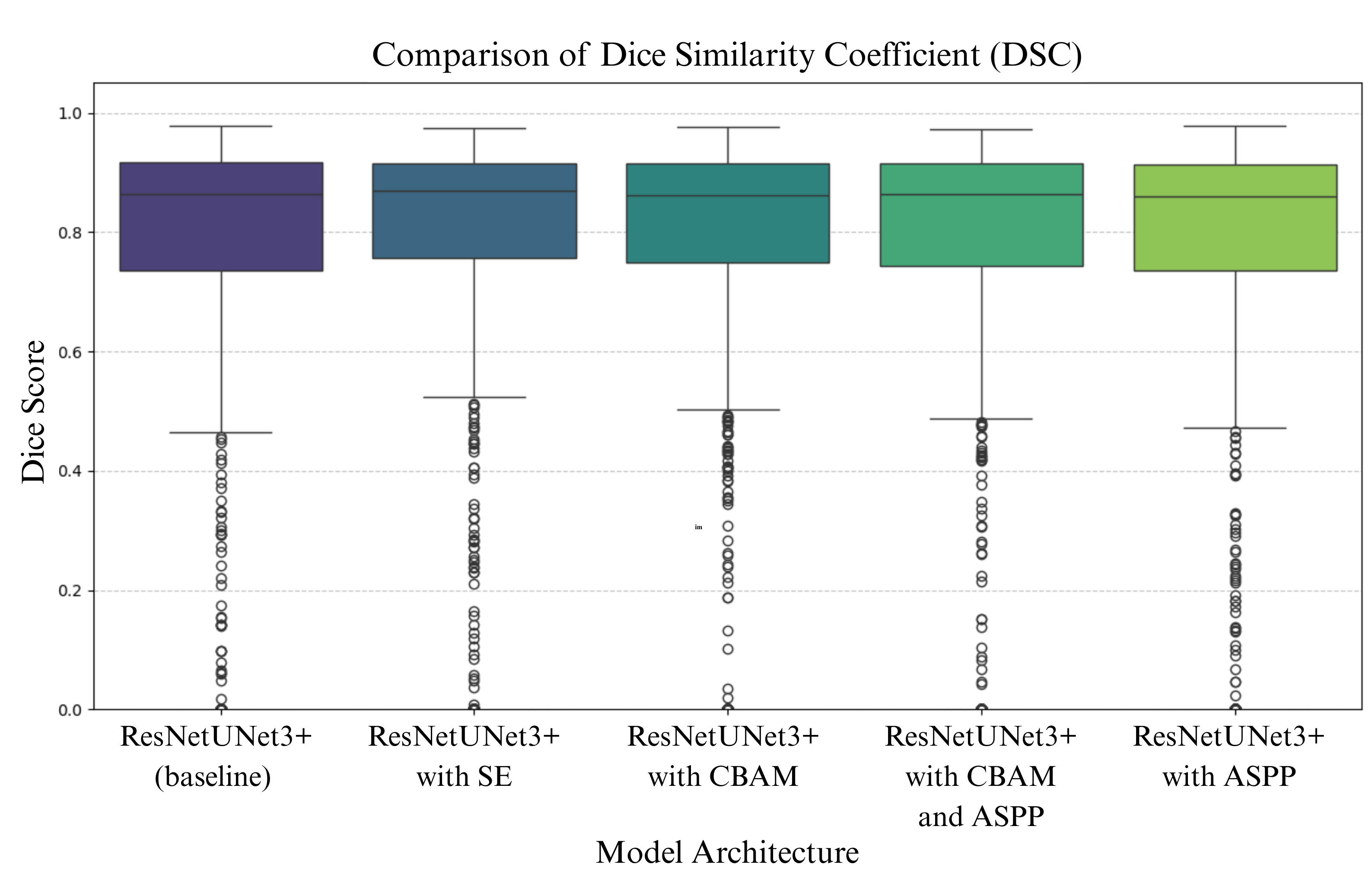}
    \caption{Box plot of Dice score of ResNetUNet3+ and its variants against ground truth on the testing set.}
    \label{fig:fig7}
\end{figure}

Table \ref{tab:table5} presents seven metrics as \(Mean \pm Standard\) Deviation. The P-value indicates the statistical significance of the Dice score difference compared to the baseline (ResNetUNet3+) using the Wilcoxon signed-rank test. Cases where the model failed to predict any tumor mask (Dice = 0) were excluded from the HD95 average to prevent undefined infinity values. 
The integration of the CBAM module yielded the most robust performance, achieving the highest nominal Dice score (\(0.755 \pm 0.268\)) and IoU (\(0.662 \pm 0.261\)). Although the improvement in the mean Dice score compared to the baseline was not statistically significant (p = 0.287), the CBAM variant demonstrated superior model stability, evidenced by the lowest standard deviation among all tested configurations (0.268 vs. 0.284 for baseline). This suggests that jointly modeling channel- and spatial-wise dependencies helps reduce performance variance, allowing the network to maintain consistent segmentation quality across diverse samples.
In contrast, the addition of ASPP alone resulted in a statistically significant degradation in performance (p < 0.001), with the mean Dice score dropping to \(0.735 \pm 0.295\). This indicates that the multi-scale context aggregation provided by ASPP may introduce noise or redundant features in this specific architecture, disrupting the precise boundary delineation required for small liver tumors. When combining ASPP with CBAM, the performance recovered (\(0.753 \pm 0.275\)), but did not surpass the CBAM-only configuration, confirming that the additional complexity of ASPP offers no distinct advantage over CBAM's feature refinement.
To further investigate the distribution of errors, we analyzed the box plots of the Dice scores (\ref{fig:fig7}). A notable observation is the presence of outlier cases where the Dice score drops to near zero across all models.

\begin{figure}[htb!] 
    \centering
    \includegraphics[width=0.8\textwidth]{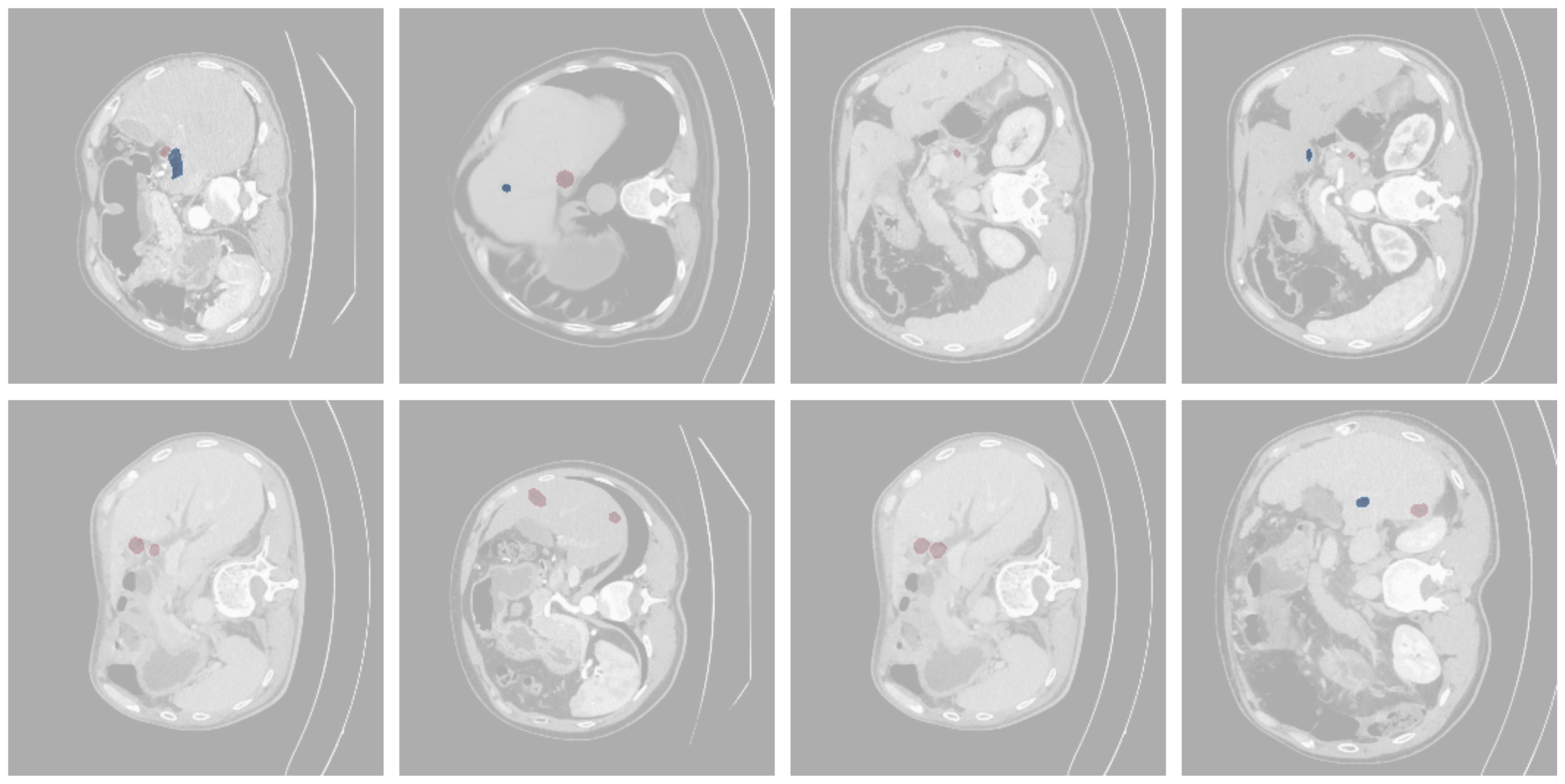}
    \caption{Visualization of hard example outliers of ResNetUNet3+ with CBAM. The predicted and groundtruth are highlighted in blue and red, respectively.}
    \label{fig:fig8}
\end{figure}

A notable observation is the presence of a long tail of outliers where the Dice score drops to near zero across all model variants. To understand the pathology behind these failures, we extracted representative samples corresponding to these outliers, as shown in Fig. \ref{fig:fig8}. These cases illustrate extremely diminutive size and disjointed (non-clustered) tumor morphologies that lead to near-zero Dice scores.

\begin{figure}[htb!] 
    \centering
    \includegraphics[width=0.8\textwidth]{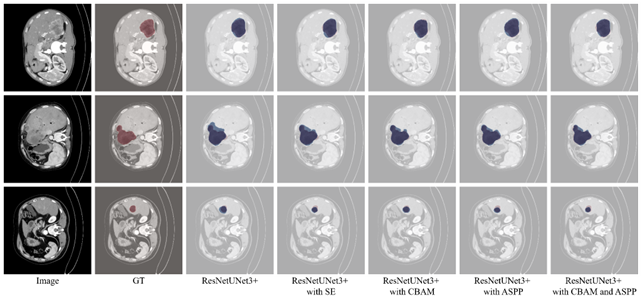}
    \caption{The visual comparison of segmentation results of ResNetUNet3+ and its variants against ground truth on the testing set.  The predicted and GT are highlighted in blue and red, respectively.}
    \label{fig:fig9}
\end{figure}

Figure \ref{fig:fig9} shows a qualitative comparison of segmentation results from the baseline model and its variants, including attention modules (SE, CBAM) and a multi-scale feature extractor (ASPP), against the ground truth on three test images. The baseline model generally identifies lesions well but sometimes produces imprecise boundaries. The CBAM variant offers improved segmentation accuracy, especially in capturing irregular lesion contours, as seen in the second row. This indicates that CBAM's spatial and channel-wise attention enhances the model's ability to differentiate lesions from healthy tissue, resulting in more precise masks.

\subsection{Discussion}
The present study investigated liver tumor segmentation on multi-phase contrast-enhanced CT images using different UNet-based architectures and attention mechanisms. The comparative evaluation highlights several important insights regarding the strengths and limitations of various model configurations, as well as their potential clinical relevance.

Experimental results demonstrated clear improvements when moving from the baseline UNet to advanced backbones. While incorporating Swin Transformer and Mamba layers enhanced feature extraction compared to the traditional UNet, ResNet-based models consistently outperformed these modern alternatives. We attribute this finding to the role of inductive biases. While Transformers and State-Space Models (Mamba) excel at modeling global dependencies, they lack the inherent translation invariance and local connectivity built into CNNs. In medical imaging regimes with limited data, these inductive biases make ResNet significantly more sample-efficient than data-hungry alternatives which typically require massive pre-training corpora to converge.

Building on this robust CNN backbone, the integration of the Convolutional Block Attention Module (CBAM) provided the most substantial performance gains. The ResNetUNet3+ with CBAM achieved the highest Dice score and IoU while minimizing boundary errors (lowest HD95). This suggests that once the backbone is stabilized by CNN priors, explicitly attention-guided feature refinement becomes the key differentiator for delineating small, low-contrast lesions.

\begin{figure}[htb!] 
    \centering
    \includegraphics[width=0.55\textwidth]{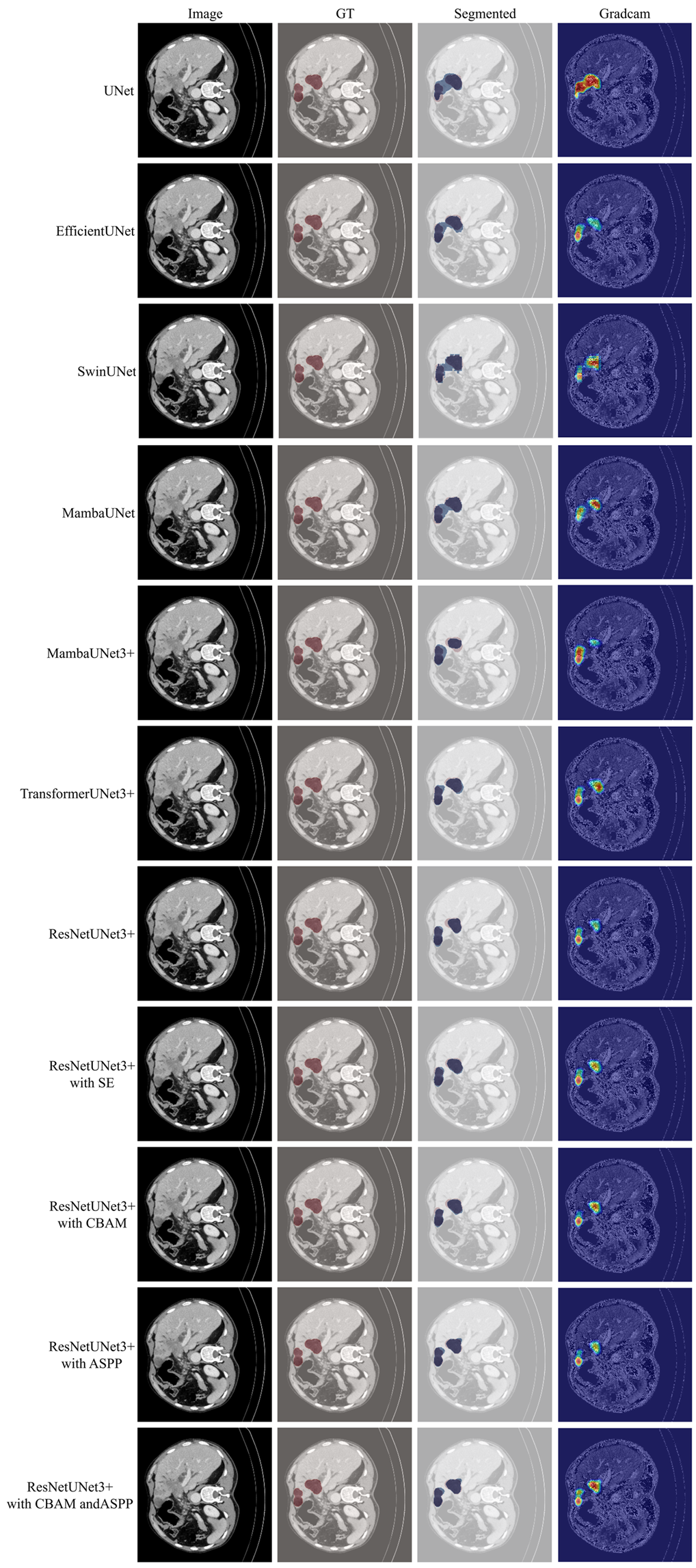}
    \caption{Sample visual explanations of all models on the testing set.}
    \label{fig:fig10}
\end{figure}

To validate these quantitative metrics, Grad-CAM visualizations (Fig. 10) were analyzed to interpret the decision-making process. These heatmaps confirmed that while weaker models often attended to irrelevant background structures leading to false positives, the proposed model consistently focused on tumor regions and their immediate context. Furthermore, Fig. \ref{fig:fig10} demonstrates that while standard UNet models tend to merge adjacent tumors into a single entity, the proposed ResNetUNet3+ with CBAM distinctively separates them, achieving boundaries that align most closely with the ground truth.

\begin{figure}[htb!] 
    \centering
    \includegraphics[width=0.8\textwidth]{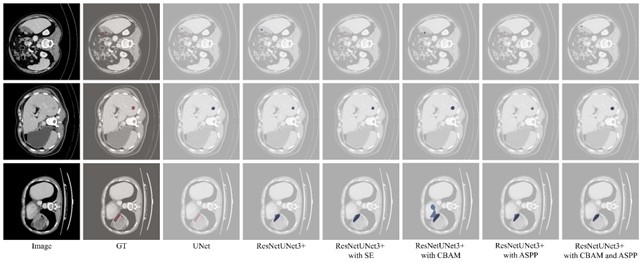}
    \caption{Some failure cases.}
    \label{fig:fig11}
\end{figure}

Although quantitative metrics indicate that ResNetUNet3+ with CBAM is the leading model, qualitative evaluation of its failure cases highlights notable shortcomings, especially regarding the segmentation of small tumors as illustrated in Fig. \ref{fig:fig11}.

While the proposed model outperforms baseline architectures, the Dice scores across all tested models reflect the inherent difficulty of the dataset. As shown in the distribution analysis (Fig. 2), the data is heavily concentrated near zero-pixel areas, presenting a distinct challenge compared to standard benchmarks. The error analysis highlights key obstacles encountered in segmenting small tumors (see Fig. \ref{fig:fig11}):
\begin{itemize}
    \item False negatives (missed tumors): The initial row of the error plot presents a scenario where a diminutive lesion exists in the ground truth. In this case, nearly all models, including the baseline, SE, and ASPP variants, fail to detect the lesion. Notably, the ResNetUNet3+ with CBAM model demonstrates partial effectiveness by detecting the approximate center of the lesion with a small mark, albeit without capturing the entire region. This outcome suggests that its attention mechanism possesses heightened sensitivity for anomaly localization, whereas other models do not recognize the lesion at all.
    \item Under-Segmentation of small lesions: The second row illustrates another challenge. All models identify the general region of the tumor; however, the baseline, SE, and ASPP variants exhibit pronounced under-segmentation, delineating only a limited portion of the lesion. By contrast, both ResNetUNet3+ with CBAM and the CBAM and ASPP variant effectively segment the lesion in its entirety. This finding highlights their superior capability to delineate the complete shape and extent of small tumors upon successful localization, thereby mitigating the under-segmentation observed with alternative approaches.
    \item Arbitrary shape of tumor regions: The third row demonstrates the segmentation error of the ResNetUNet3+ model with CBAM when addressing tumors with non-circular, irregular shapes. Compared to other methods such as ResNetUNet3+ and its variants with SE and ASPP, the proposed model tends to segment a larger region. Notably, ResNetUNet3+ with both CBAM and ASPP achieves the highest performance by accurately delineating the tumor region. This observation accounts for the highest Precision score attained by the ResNetUNet3+ with CBAM and ASPP, as reported in Table \ref{tab:table5}.
\end{itemize}

In summary, although the ResNetUNet3+ with CBAM model achieves the highest overall performance, the intrinsic complexity of the dataset - dominated by diminutive and irregular lesions - imposes a performance ceiling (\(Dice \sim 0.755\)). As illustrated in Fig. \ref{fig:fig11}, failures manifest in distinct patterns: extremely small lesions often result in missed detections (Row 1), though the CBAM module demonstrates superior sensitivity compared to baselines. Furthermore, while less advanced models suffer from significant under-segmentation (Row 2), the CBAM-equipped models consistently improve boundary delineation. However, this heightened sensitivity involves a trade-off, occasionally resulting in false positives for non-circular morphologies (Row 3). Consequently, the reported metrics reflect the specific difficulty of delineating these pathological subtypes rather than architectural deficiencies.

\subsection{Limitation}
Despite the encouraging segmentation results, several limitations warrant consideration and outline directions for future research. First, the current study utilizes a dataset from a single institution. While this ensures consistent data quality, it limits the evaluation of the model's generalizability across different scanners, imaging protocols, and pathological subtypes. The current performance metrics may reflect adaptation to this specific data distribution, and validation on multi-center datasets will be required to ensure robustness.

Second, the adopted approach decomposed 3D volumetric data into 2D slices to optimize computational efficiency. This strategy inherently leads to a loss of inter-slice spatial correlations and longitudinal connectivity, which are critical for fully leveraging multi-phase CECT information. Future studies are expected to extend this architecture to 3D volumetric segmentation to better capture these cross-dimensional features.

Third, although the attention mechanisms employed in this study demonstrated effectiveness, the field is rapidly evolving. More recent state-of-the-art methods, such as the Parameter-Shared Encoder (PSE) \cite{wang2025pse}, will be evaluated in subsequent work to further enhance feature representation.

Finally, although Grad-CAM analysis provided theoretical insights into the model's focus on tumor boundaries, the structural generalizability of the specific module combinations has only been demonstrated on the current dataset. Extensive testing on diverse cohorts is required to fully establish the theoretical and practical superiority of this combination.

\section{Conclusion}
\label{sec:conclusion}
The study explored liver tumor segmentation on CT images using UNet- and UNet3+-based architectures with diverse backbone configurations. Through comprehensive experiments, it was shown that while modern Transformer- and Mamba-based backbones improved feature representation, the combination of UNet3+ with a ResNet backbone and CBAM attention offered the most balanced performance. Although the inherent difficulty of segmenting diminutive lesions resulted in high variance across all models, the proposed architecture mitigated this instability, achieving a lower standard deviation compared to the baseline (0.268 versus 0.284). Consequently, the model delivered competitive Dice and IoU scores while reducing boundary errors, indicating improved reliability in handling the hard cases where standard architectures frequently fail.

Beyond quantitative improvements, the inclusion of Grad-CAM visualizations improved the interpretability of the model, highlighting its ability to focus on clinically relevant tumor regions. These results underscore the potential of attention-augmented UNet3+ models in advancing computer-aided liver tumor detection and segmentation, which can contribute to more accurate diagnosis and treatment planning.

\section*{Declaration of conflicting interest}
The authors declared no potential conflicts of interest with respect to the research, authorship, and/or publication of this article.

\section*{Funding}
This study was supported by Vietnam National University Ho Chi Minh City (VNU-HCM) under grant DS2023-28-02. The funder had no role in study design, data collection and analysis, decision to publish or manuscript preparation.

\section*{Data availability statement}
Data supporting the findings of this article are publicly available at \url{https://doi.org/10.57760/sciencedb.12207} \cite{lou2025comprehensive}.

\section*{ORCID iD}
Doan-Van-Anh Ly: \url{https://orcid.org/0009-0006-6898-0545}\\
Thanh-Hai Le: \url{https://orcid.org/0000-0002-3212-3940}\\
Thi-Thu-Hien Pham: \url{https://orcid.org/0000-0001-5808-3214}

\bibliographystyle{unsrt}  
\bibliography{references}  

\end{document}